\documentclass[runningheads]{llncs}
\usepackage{graphicx}
\usepackage{cite}
\usepackage{algorithm}
\usepackage{algorithmic}
\usepackage{amsmath, amsmath, amssymb}
\usepackage{mathrsfs}
\usepackage{bm}
\usepackage{makecell}
\usepackage{multirow}

\usepackage{color}

\usepackage[colorlinks=true, linkcolor=blue, anchorcolor=blue, citecolor=blue]{hyperref}

\newcommand{\etal}{\mbox{\emph{et al.}}}

\newcommand{\tablestyle}[2]{\setlength{\tabcolsep}{#1}\renewcommand{\arraystretch}{#2}\centering\footnotesize}

\begin{document}
\title{End-to-end cell recognition by point annotation\thanks{These authors contributed equally to this work.}}

\author{Zhongyi Shui\inst{1,2,*} \and 
Shichuan Zhang\inst{1,2,*}\and
Chenglu Zhu\inst{1,2}\and
Bingchuan Wang\inst{3}\and
Pingyi Chen\inst{1,2}\and
Sunyi Zheng\inst{1,2}\and
Lin Yang\inst{2}}

\authorrunning{Z. Shui et al.}
\institute{College of Computer Science and Technology, Zhejiang University \and
School of Engineering, Westlake University \and
School of Automation, Central South University \\
\email{yanglin@westlake.edu.cn}}

\maketitle              
\begin{abstract}
Reliable quantitative analysis of immunohistochemical staining images requires accurate and robust cell detection and classification. Recent weakly-supervised methods usually estimate probability density maps for cell recognition. However, in dense cell scenarios, their performance can be limited by pre- and post-processing as it is impossible to find a universal parameter setting. In this paper, we introduce an end-to-end framework that applies direct regression and classification for preset anchor points. Specifically, we propose a pyramidal feature aggregation strategy to combine low-level features and high-level semantics simultaneously, which provides accurate cell recognition for our purely point-based model. In addition, an optimized cost function is designed to adapt our multi-task learning framework by matching ground truth and predicted points. The experimental results demonstrate the superior accuracy and efficiency of the proposed method, which reveals the high potentiality in assisting pathologist assessments.

\end{abstract}

\section{Introduction}
Quantitative immunohistochemistry image analysis is of great importance for treatment selection and prognosis in clinical practice. Assessing PD-L1 expression on tumor cells by the Tumor Proportion Score (TPS) \cite{doroshow2021pd} is a typical example. However, large inter-reader variability may lead to a lack of consistency in the assessment, and the labor-intensive manual observation cannot guarantee the accuracy of clinical diagnosis. Hence, there is a strong demand to develop an AI tool that provides high-quality quantification results through accurate cell detection and classification.

With the advances in computing power and increased data volume in recent years, deep learning-based cell recognition methods \cite{zhou2018sfcn,qu2019weakly,chamanzar2020weakly} have shown more advantages over conventional methods. However, deep learning models usually require a massive amount of labeled data. To improve the labeling efficiency, pixel-wise annotations are gradually replaced with point annotations in cell recognition. Recent studies usually build a regression model to predict probability density maps (PDMs) of cells. 
In this case, pre-processing is required to convert point annotations into reference density maps (RDMs) for model training. 
Many studies on RDMs generation have been conducted to improve the cell recognition performance of the regression models. For instance, Zhou \etal \cite{zhou2018sfcn} convolved raw point labels with a customized Gaussian kernel to generate circle-shape masks. Qu \etal \cite{qu2019weakly} utilized Voronoi transformation and local pixel clustering methods to generate pseudo labels. Liang \etal \cite{liang2019enhanced} proposed a repel coding method to suppress the responses between two cell centers. Conversely, the post-processing is required to obtain the final recognition results from the predicted PDMs of cells, which is also a challenge for the identification of adjacent cells. The local maximum searching algorithm \cite{huang2020bcdata} is typically adopted to locate cells. However, it is difficult to find an appropriate set of hyper-parameters to deal with the scale variations between tissues and cells in the immunohistochemical (IHC) membrane-stained image.

To address the aforementioned issues, we develop an end-to-end model that regresses and classifies preset anchor points to recognize cells directly. In this point-based model, we combine the high-level semantic features with low-level local details, which benefit cell localization and classification \cite{cosatto2008grading,veta2013automatic,lin2017feature} in large-scale histopathological scenes. Moreover, an improved cost function is proposed towards multi-task learning to determine which ground truth point should the current proposal point be responsible for in the one-to-one matching procedure \cite{song2021rethinking}. In addition, to prevent the model from overfitting to noisy labels, we extend the classification loss with an L2 normalization term to penalize sharp softmax predictions. The experimental results demonstrate the superior performance and efficiency of the proposed method for dense cell recognition over previous studies.

\section{Methods}
The overall framework of the proposed model is depicted in Fig.~\pageref{fig:framework}, which includes a feature extraction module, a multi-task prediction module, and a proposal matching module. Initially, the feature extraction module generates multi-level representation from the encoding layers by pyramidal feature aggregation. Subsequently, the generated representation is performed separately in three branches of regression, detection, and classification, which can further improve the cell recognition performance by multi-task learning. Finally, the proposal matching module matches the proposal points to the ground truth targets.
\begin{figure}[t]\setlength{\abovecaptionskip}{0pt}
	\setlength{\belowcaptionskip}{0pt}
	\centering
	\includegraphics[width=\linewidth]{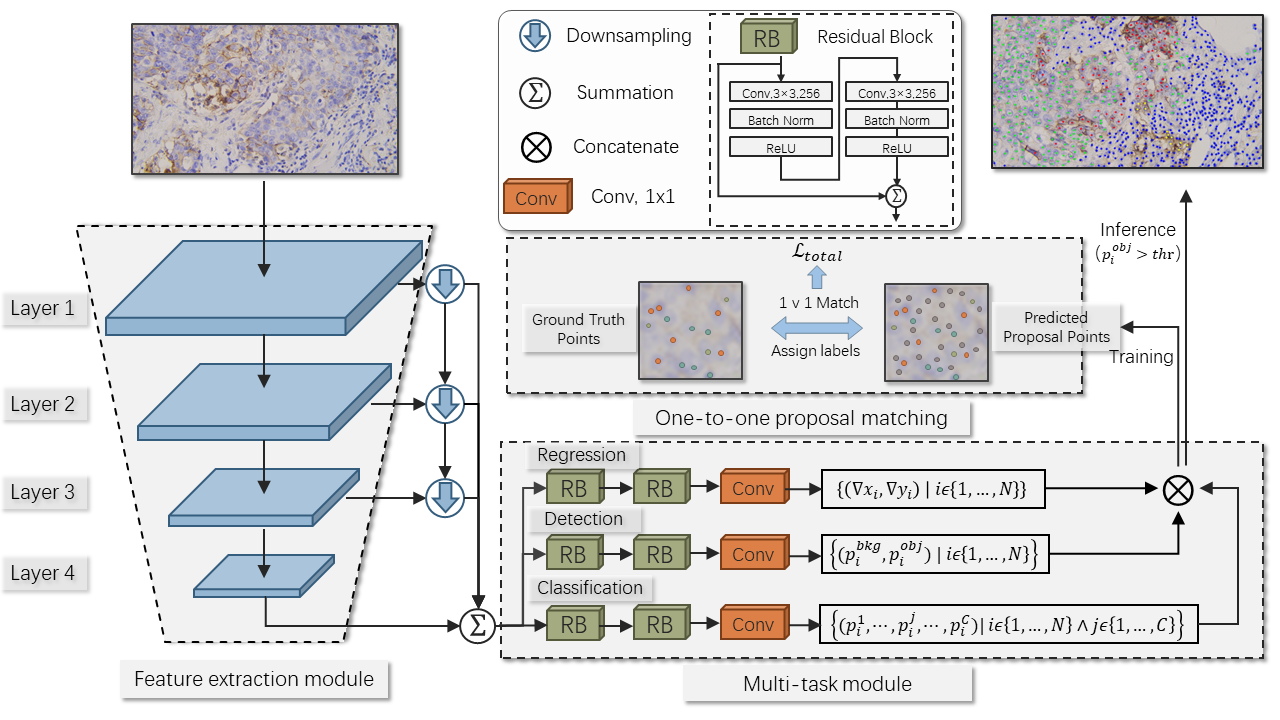}\\
	\caption{Framework of our model}\label{fig:framework}
\end{figure}

\paragraph{\textbf{Pyramidal Feature Aggregation.}}\label{sec:PFA}
The high-dimensional semantic representation can usually be obtained by the encoding of a deep network model. Specifically, the semantic information of an image is gradually extracted from low-level to high-level by a series of layers and then used for downstream tasks to complete the model training. However, it may not be sufficient to recognize each cell with the high-level semantic generalization in a large-scale histopathological scene. The low-level regional descriptions (color, edge, etc.) are also considered to promote cell recognition \cite{cosatto2008grading,veta2013automatic}. Consequently, pyramidal features extracted from all layers are aggregated with a unified shape to generate adequate representation.

\paragraph{\textbf{Multi-task Learning.}}
The success of some previous studies \cite{zhang2022weakly, chamanzar2020weakly} of cell recognition shows the advantage of the multi-task strategy in weakly supervised learning. Thus a similar structure is introduced in our recognition framework.
Instead of general recognition models which need to recover the prediction to the original image size, the high-dimensional features are directly fed into three separate task modules (regression, detection, and classification) in our method.
Specifically, two residual blocks followed by a $1\times1$ convolutional layer comprise each task module. 
The outputs of the regression head are the offsets $(\nabla x_i, \nabla y_i)$ of each proposal point $p_i$, which refines the cell location prediction as follows:
\begin{equation}
\hat{x}_i = x_i + \nabla x_i,\quad \hat{y}_i = y_i + \nabla y_i
\end{equation}\label{eq:location}in which $(x_i, y_i)$ is the initial coordinate of $p_i$ and $(\hat{x}_i, \hat{y}_i)$ represents its refined coordinate. The detection head outputs the scores $(p_i^{bkg},p_i^{obj})$, where $p_i^{obj}$ indicates its confidence of being a cell center. Lastly, the classification head outputs the category scores $\bm{p_i}=(p_i^1,\cdots,p_i^{C})$, in which $p_i^j$ indicates its probability of being the $j$-th cell category and $C$ is the number of cell categories. All output information will be combined for training and inference.

\paragraph{\textbf{Proposal Matching.}}
Generally, cell recognition aims at predicting point sets of indeterminate size is inherently an open set problem \cite{xiong2019open}. The key is to determine the unique matching between the current prediction and various ground truth points.
To improve optimization efficiency and suppress duplicate predictions, the one-to-one matching strategy described in the study of Song \etal \cite{song2021rethinking} is introduced. To be specific, based on the optimized cost function that considers distance, target confidence, and category confidence simultaneously, a pair-wise cost matrix is first constructed as follows:
\begin{equation}
\begin{aligned}
\mathcal{E}=\alpha \mathcal{D}-\mathcal{F}_{t}-\mathcal{F}_{c}= (\alpha \left \| \bm{p_i^{loc}} - \bm{p_j^{loc*}} \right \|_{2} - p_{i}^{obj} - p_{i}^{c_j^*})_{i=1,\cdots,M,\; j=1,\cdots,N}
\end{aligned}
\end{equation}
where $\bm{p_i^{loc}}=(\hat{x}_i, \hat{y}_i)$ and $\bm{p_j^{loc*}}=(x_i^*, y_i^*)$ represents the coordinates of the $i$-th proposal point and the $j$-th ground truth point, $c_j^*$ is the category of the $j$-th ground truth point and $\alpha$ is a coefficient used to balance the weights of the distance factor and the confidence factors. Note that $M \ge N$ should be ensured to produce enough predictions. Finally, the Hungarian algorithm \cite{kuhn1955hungarian, song2021rethinking} is conducted on $\mathcal{E}$ to find the minimum-cost matching, the result of which can be represented with $\left\{(p_{\delta(j)}, p_j^*),j=1,\cdots,N\right\}$, where $\delta(\cdot)$ is the mapping function from the subscript of one ground point to that of its matching proposal point. The predicted points that fail to match any ground truth point are treated as negative proposals.

\paragraph{\textbf{Loss Function.}} 
We adopt the mean square error (MSE) loss for the regression training\cite{song2021rethinking}. 
\begin{equation}
\mathcal{L}_{reg} = \frac{1}{N}\sum_{j=1}^{N}||\bm{p_{\delta(j)}^{loc}}-\bm{p_j^{loc*}}||_2 \\
\end{equation}
in which $\beta$ is a balance factor.

For the cell detection task, the labels are noise-free because there is a high probability that cells exist at the annotated points. Therefore, the cross entropy (CE) loss is adopted to prompt the model to focus on the samples whose softmax predictions are less consistent with provided labels \cite{zhang2018generalized}.
\begin{equation}
\mathcal{L}_{det} = -\frac{1}{M}\left(\sum_{i\in\mathcal{P}}\log(p_i^{obj}) + \beta\sum_{{i\in\mathcal{N}}}\log(p_i^{bkg})\right)
\end{equation} 
where $\mathcal{P}=\{\delta(j) \mid j=1,\cdots,N\}$ is the set of positive proposals and $\mathcal{N}=\{1,\cdots,M\} \setminus \mathcal{P}$ is the set of negative proposals.

For the cell classification task, the categorical labels are noisy due to intra-reader variability. Therefore, to improve the robustness of our model to label noise during training, we extend the generalized cross entropy (GCE) loss \cite{zhang2018generalized} with an L2 regularization term as follows:
\begin{equation}\label{eq4}
\widetilde{\mathcal{L}}_q(\bm{p_{\delta(j)}}, c_j^*) = \frac{1-(p_{\delta(j)}^{c_j^*})^q}{q}+\gamma||\bm{p_{\delta(j)}}||_2
\end{equation} 
where $q$ is a parameter to balance convergence and noise robustness of the GCE loss.
The additional term prevents the model from being overconfident and improves its generalization ability. The total classification loss is:
\begin{equation} 
\mathcal{L}_{cls} = \sum_{j=1}^{N}\widetilde{\mathcal{L}}_q(\bm{p_{\delta(j)}}, c_j^*)
\end{equation}
Finally, the total loss is as follows:
\begin{equation}
\mathcal{L}_{total} = \lambda\mathcal{L}_{reg}+\mathcal{L}_{det}+\mathcal{L}_{cls}
\end{equation}
where $\lambda$ adjusts the weight of the regression loss.

\section{Experiments}

\subsection{Dataset description and experimental settings}

\paragraph{\textbf{Dataset.}} 
To evaluate the proposed method, we performed experiments on PD-L1 IHC stained images of tumor tissues, which are utilized to analyze the PD-L1 expression in non-small cell lung cancer samples. We first collected 485 patches with the resolution of $1920\times1080$ from the desensitized data obtained at $40\times$ magniﬁcation of whole slide images. Subsequently, three pathologists labeled all cells in the patches by point annotation method with four categories including PD-L1 positive tumor cells, PD-L1 negative tumor cells, PD-L1 positive non-tumor cells, and PD-L1 negative non-tumor cells. Finally, the patches are randomly split into the training set and test set at a ratio of 4:1. Notably, to guarantee the quality of the test set, a senior pathologist further verified the annotations of test images.

\paragraph{\textbf{Implementation Details.}}
We compared our method with seven competitors under the same training and test environment. To be specific, for the training of regression models, we applied a 2D Gaussian ﬁlter on the point annotations to generate RDMs, which ensures the highest response at the center point of each cell. The optimal parameters (kernel size $7\times7$, $sigma=6$) were selected through experiments. Note that all regression models are trained using the binary cross entropy (BCE) loss and intersection over union (IOU) loss, whose weights are set to 0.8 and 0.2, respectively.

Since cells are highly dense in the constructed dataset, we preset 5 anchor points in each $32\times32$ pixel region on the original image. One is in the center and the other four are generated by shifting the center point with (-8,-8), (-8,8), (8,8) and (8,-8) pixels respectively. The other hyper-parameters of our method are set as follows, $\alpha=0.05, \beta=0.6, \gamma=0.1, q=0.4$ and $\lambda=2\times10^{-3}$. Data augmentation including random resized cropping and flipping are used in the training stage.

We trained all networks on a single NVIDIA A100-SXM4 GPU. The AdamW optimizer \cite{loshchilov2017decoupled} was used for minimizing the loss function with an initial learning rate of $10^{-4}$ and weight decay of $10^{-4}$.

\paragraph{\textbf{Evaluation Metric.}}
To evaluate the performance of different methods, we use precision (P), recall (R) and F1 scores that are calculated as follows:
\begin{equation}
\rm P=\frac{TP}{TP+FP},R=\frac{TP}{TP+FN},F1=\frac{2*P*R}{P+R}
\end{equation}
where TP, FN and FP represent the number of true positives, false negatives and false positives,  which are counted through the quantitative evaluation strategy adopted by Cai \etal \cite{cai2021generalizing}. Note that the radius of the valid matching area was set to 12 pixels in this work. In addition, the average inference time on test patches was introduced as an additional indicator for evaluating the analysis efficiency, the unit of which is seconds.

\subsection{Experimental results}
The quantitative comparison results are listed by cell detection and classification as Table.\ref{tab:cmp} shows.
\begin{table*}[h]
  \centering
  \tablestyle{4pt}{1.4}
  \scriptsize
  \caption{Comparison of cell recognition with diﬀerent methods}
  \resizebox{\textwidth}{!}{\begin{tabular}{c|c|ccc|ccc|c}
  \multirow{2}{*}{Type} & \multirow{2}{*}{Model} & \multicolumn{3}{c|}{Detection} & \multicolumn{3}{c|}{Classification} & \multirow{2}{*}{Time} \\
  \cline{3-8}
  ~& ~ & P & R & F1 & P & R & F1 & (seconds) \\
  \hline
  \multirow{5}{*}{Single task}
  ~ & FCRN-A \cite{xie2018microscopy} & 79.67 & 70.67 & 74.90 & 55.76 & 47.68 & 50.91 & 2.31 \\
  ~ & FCRN-B \cite{xie2018microscopy} & 68.29 & 78.27 & 72.94 & 47.46 & 52.04 & 49.30 & 3.04 \\
  ~ & UNet \cite{falk2019u} & 82.90 & 71.70 & 76.89 & 63.76 & 49.98 & 54.60 & 2.46 \\
  ~ & ResUNet \cite{zhang2018road} & 83.49 & 70.02 & 76.17 & 56.66 & 42.99 & 46.59 & 2.56 \\
  ~ & DeepLab v3+ \cite{chen2018encoder} & 77.64 & 78.70 & 78.17 & 64.45 & 60.70 & 61.75 & 2.51 \\
  \hline
  \multirow{3}{*}{Multi-task}
  ~ & Zhang \cite{zhang2022weakly} & 75.89 & 79.10 & 77.46 & 63.48 & 64.80 & 63.86 & 2.52 \\
  ~ & P2PNet \cite{song2021rethinking} & 80.34 & 82.91 & 81.61 & 67.33 & 67.85 & 67.30 & 0.12 \\
  ~ & \textbf{ours} & \textbf{81.21} & \textbf{85.17} & \textbf{83.14} & \textbf{68.13} & \textbf{70.36} & \textbf{69.09} & \textbf{0.11} \\
  \end{tabular}}
\label{tab:cmp}
\end{table*}

\begin{figure}[!h]
    \centering
    \includegraphics[width=\linewidth]{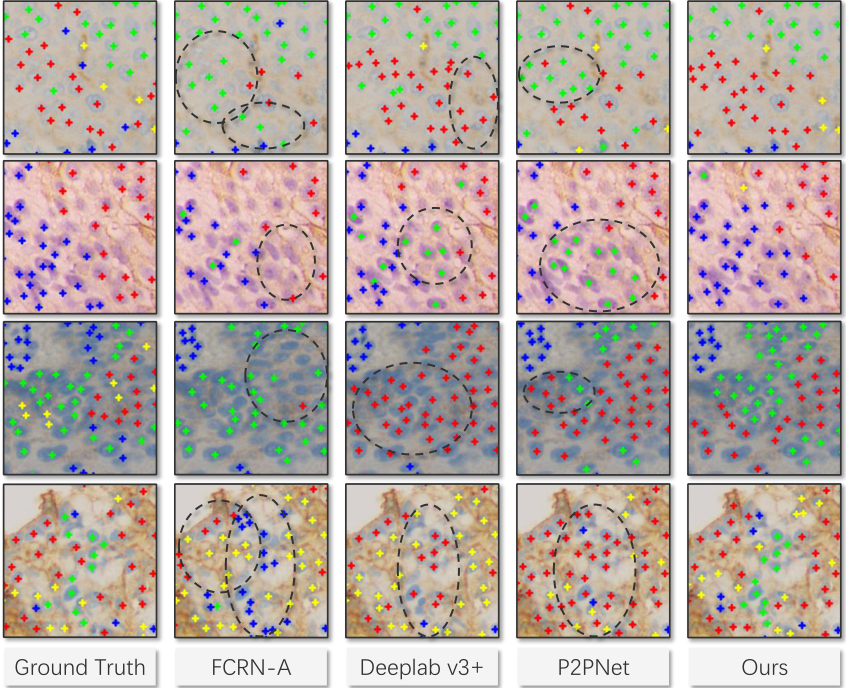}
    \caption{Visualization results of different methods. We mark positive tumor cells, negative tumor cells, positive non-tumor cells, and negative non-tumor cells in red, green, yellow, and pink, respectively.}
    \label{fig:vis}
\end{figure}

In comparison with the model proposed by Zhang \etal \cite{zhang2022weakly} that performs best among previous cell recognition studies, our model improves the F1 score by $5.68$ percent point in cell detection and $5.23$ percent point in cell classification, respectively. 
In addition, the average inference time of the proposed model is nearly $23$ times shorter than that of \cite{zhang2022weakly}, which demonstrates its higher practicality as quantitative TPS calculation requires a wide range of cell recognition with hundreds of patches cropped from a whole-slide image. The superiority of our method is intuitively reflected in the visualization results shown in Fig.~\ref{fig:vis}, where the black dashed area indicates the inadequacy of competitors.

The regression models, which recognize cells by predicting the PDMs, may yield missed detections due to required post-processing.
Specifically, if the minimum distance between peaks is set to a small value in the local maximum searching algorithm, overlapping cells are easily mistaken as a connected whole. 
On the contrary, if a larger value is set, missed detections may be caused by filtering out cells with weaker intensity.
Therefore, predicting points directly can avoid the performance degradation brought by post-processing.

\begin{figure}[t]
    \centering
    \setlength\tabcolsep{4pt}
    \begin{tabular}{c}
    \includegraphics[width=0.9\linewidth]{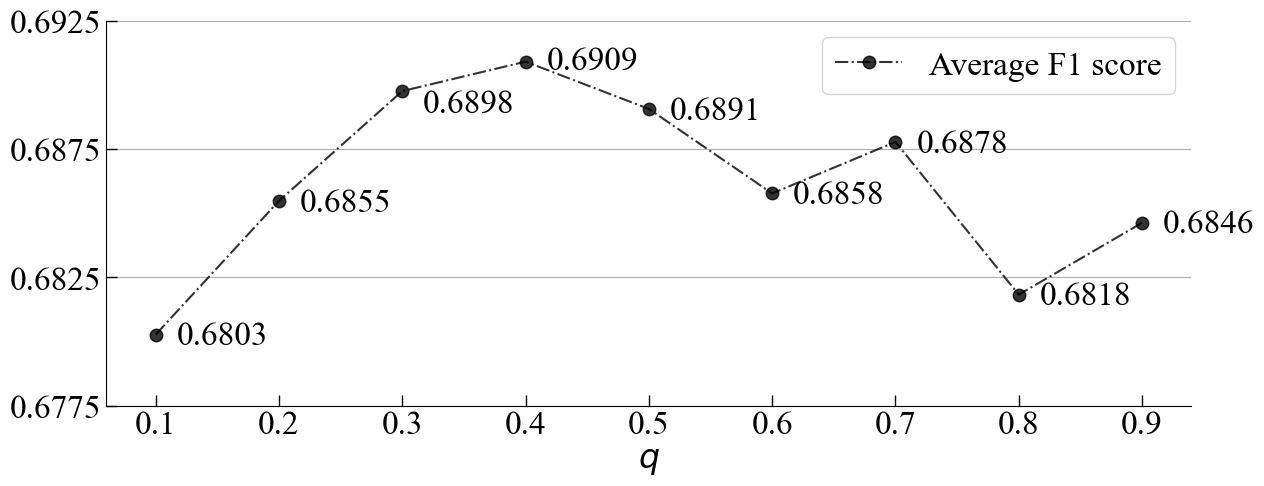}
    \end{tabular}\vspace{0mm}
    \caption{The cell classification performance of our model with $q$ ranging from $0.1$ to $0.9$}
    \label{fig:hyper-para}
\end{figure}

We also recorded the average F1 score of cell classification with $q$ ranging from $0.1$ to $0.9$ to clarify the impact of label noise on the cell recognition performance of our model in Fig.~\ref{fig:hyper-para}. To be specific, the GCE loss can be considered as the CE loss when $q\rightarrow0$ and the MAE loss when $q\rightarrow1$. The CE loss is sensitive to label noise but has good convergence characteristics. On the contrary, the MAE loss is robust to label noise while challenging to optimize. As shown in Fig.~\ref{fig:hyper-para}, the average F1 score of classification is the lowest when $q=0.1$. With the increase of $q$ in the GCE loss, our model becomes more robust to noisy labels, and its performance gradually improves. The highest F1 score is obtained when $q=0.4$. As $q$ increases further, the performance of our model degrades because the convergence of the GCE loss gets worse.

\paragraph{\textbf{Ablation Study}}
Ablation experiments are built to assess the improvement for the baseline (P2PNet) as Table.~\ref{tab:ablation} shows.
\begin{table*}[h]
  \centering
  \tablestyle{4pt}{1.3}
  \scriptsize
  \caption{Pipeline component analysis of our proposed method}
  \begin{tabular}{l|ccc|ccc}
  \multirow{2}{*}{\textbf{Method}} & \multicolumn{3}{c|}{Detection} & \multicolumn{3}{c}{Classification}\\
  \cline{2-7}
  ~ & P & R & F1 & P & R & F1 \\
  \hline
  Baseline & 80.34 & 82.91 & 81.61 & 67.33 & 67.85 & 67.30 \\
  Baseline+PFA & \textbf{81.98} & 84.12 & 83.03 & 68.95 & 68.93 & 68.41 \\
  Baseline+PFA+IC & 81.21 & \textbf{85.17} & \textbf{83.14} & \textbf{68.13} & \textbf{70.36} & \textbf{69.09} \\
  \end{tabular}
\label{tab:ablation}
\end{table*} 
It is worth noticing that PFA improves the F1 score by $1.42$ percent point in detection and $1.11$ percent point in classification, respectively. The improvement demonstrates the effectiveness of using low-level features to promote cell recognition performance. In addition, the designed independent classification (IC) branch and the noise-robust GCE loss can further improve the performance of the model.

\section{Conclusion}
In this paper, we propose a purely point-based cell recognition model that can recognize cells in an end-to-end manner. Unlike the mainstream regression-based methods that require pre-processing to generate reference density maps and post-processing of local maximum searching to locate cells, the proposed model can directly recognize cells by preset anchor points. Specifically, we propose a pyramidal feature aggregation strategy to combine semantic features with detailed information, which helps the model distinguish the boundaries and categories of cells. In addition, an optimized cost function is proposed to adapt our multi-task learning framework by matching ground truth and proposal points. An extended loss function is also proposed to improve the robustness of our model to noisy labels. Finally, the experimental results demonstrate the superior performance and efficiency of our proposed method, and the results of the ablation study show the effectiveness of our designed modules. Such a model can reduce the workload of pathologists in the quantitative analysis of IHC stained images.

\noindent\textbf{Acknowledgements.} This work was funded by China Postdoctoral Science Foundation (2021M702922).

%
%
\bibliographystyle{splncs04}
\bibliography{mybib}

\end{document}